\documentclass{article}

\usepackage{times}
\usepackage{graphicx}
\usepackage{subfigure} 

\usepackage{natbib}

\usepackage{algorithm}
\usepackage{algorithmic}

\usepackage{amsmath, amsfonts, amssymb, amsthm, amstext}

\usepackage{hyperref}

\usepackage[accepted]{icml2015}

\newcommand{\C}{\mathbb{C}}
\newcommand{\LL}{\mathrm{L}}

\newcommand{\R}{\mathbb{R}}

\newcommand{\Z}{\mathbb{Z}}

\icmltitlerunning{Quantum Energy Regression using Scattering Transforms}

\begin{document} 

\twocolumn[
\icmltitle{Quantum Energy Regression using Scattering Transforms}

\icmlauthor{Matthew Hirn}{matthew.hirn@ens.fr}
\icmladdress{\'Ecole Normale Sup\'erieure, D\'epartement d'Informatique, 45 rue d'Ulm, 75005 Paris, France}
\icmlauthor{Nicolas Poilvert}{nup18@psu.edu}
\icmladdress{Millennium Science Complex, The Pennsylvania State University, University Park, PA 16801 USA}
\icmlauthor{St\'ephane Mallat}{stephane.mallat@ens.fr}
\icmladdress{\'Ecole Normale Sup\'erieure, D\'epartement d'Informatique, 45 rue d'Ulm, 75005 Paris, France}

\icmlkeywords{Quantum Energy Regression Scattering Transforms}

\vskip 0.3in
]

\begin{abstract}
We present a novel approach to the regression of quantum mechanical energies based on a scattering transform of an intermediate electron density representation. A scattering transform is a deep convolution network computed with a cascade of multiscale wavelet transforms.
It possesses appropriate invariant and stability properties for quantum energy regression.
This new framework 
removes fundamental limitations of Coulomb matrix based energy regressions, and numerical
experiments give state-of-the-art accuracy over planar organic molecules. 
\end{abstract}

\section{Introduction}
\label{introduction}

Estimating the ground state energy of atoms and molecules is one of the most fundamental and studied topics in computational quantum mechanics. The traditional approach to this problem has been to devise clever ways to solve for the equations of quantum mechanics. Recently though, it has been proposed to
attack the problem from a Machine Learning (ML) perspective \cite{Rupp2012}.

Most machine learning approaches are representing the molecular state as a Coulomb matrix of pairwise energy terms \cite{Rupp2012,Hansen2013}. An important limitation of a
Coulomb representation is that it depends on an ordering of the atoms in the molecule. When the atom ordering is changed, the Coulomb matrix changes while the energy does not.

A first contribution of this paper is to introduce a new molecular representation in the form of a two or three dimensional signal. We define a one-to-one mapping between the molecular state and a real-valued positive function defined over $\R^2$ or $\R^3$, 
which has the physical interpretation of an approximate electron density. This first step circumvents the issue of atom ordering.
In numerical applications, we restrict ourselves to \textit{planar} molecules, with atoms lying on the same molecular plane. We will therefore use two-dimensional electron densities, but three dimensional extensions are calculated similarly.

Regression of a high dimensional functional requires the use of prior knowledge
of its invariance and stability properties.
For quantum energy regression, many invariance and stability properties of the energy function are known. Imposing these same properties onto a representation is important to obtain accurate regressions.
Scattering transforms introduced by \cite{Mallat2012} are examples
of convolutional networks \cite{Krizhevsky2012,Sermanet2013}, computed with iterated wavelet transforms, which yield appropriate invariants.
Our second contribution shows 
that these scattering transforms, successfully used for image classification 
\cite{MallatBruna2013,oyallon:scatObjectClass2014}, 
can be used to regress quantum energies to state-of-the-art accuracy.

The paper is organized as follows. Section \ref{molecular_representations} discusses the properties of "good" molecular representations, and presents the current state-of-the-art along with its known limitations. Section \ref{energy_regression} introduces the problem of regressing energies through sparse linear expansions over dictionaries of density functionals. Section \ref{dictaser} gives the mathematical details of a number of invariant representations used in this work. Section \ref{sec: numerical experiments} describes the setup of the numerical experiments along with the values of all the numerical parameters used in generating the dictionaries in \ref{dictaser}. Finally, Section \ref{results} analyses the results of our experiments.

\section{Molecular representations for energy regression}
\label{molecular_representations}

High dimensional regressions must take advantage of prior information and invariances of the function that is estimated in order to reduce the problem dimensionality.

We start by outlining properties that should be satisfied by a 
molecular representation in an energy regression context. Next, the current best-in-class representation based on Coulomb matrices will be presented, along with its known limitations.

\subsection{Desirable properties for a molecular representation}
\label{desirable_properties_of_representations}

We are interested in regressing molecular atomization energies. A molecule containing $K$ atoms is entirely defined by its nuclear charges $z_{k}$ and its nuclear position vectors $p_{k}$ indexed by $k$. Denoting by $x$ the state vector of a molecule, we have
\begin{equation*}
x = \{ (p_k, z_k) \in \R^3 \times \R : k = 1, \ldots, K \}.
\end{equation*}
Since the target value that we are trying to regress is a scalar representing a physical energy, we know that:
\begin{description}
\item [Permutation invariance] The energy is invariant to the permutation of the indexation of atoms in the molecule.
\item [Isometry invariance] The energy is invariant to translations,
  rotations, and symmetries of the molecule and hence to any orthogonal operator. 
\item [Deformation stability] The energy is differentiable with respect to the distances between atoms.
\item [Multiscale interactions] The energy has a multiscale structure, with highly energetic covalent bonds between neighboring atoms, and weaker energetic exchanges at larger distances, such as Van Der Waals interactions.
\end{description}
The deformation stability stems from the fact that a small deformation of the molecule induces a small modification of its energy. The primary difficulty is to construct a representation 
which satisfies these four properties, while simultaneously containing a rich enough set of descriptors to accurately regress the atomization energy of a diverse collection of molecules.  

\subsection{Coulomb matrix representation}

Representations of distributions of points, which are invariant to
orthogonal transformations and stable to deformations can be defined from pairwise
distances between these points. This is the strategy adopted by
 the current state-of-the-art in molecular energy regression, which makes use of a so-called \textit{Coulomb matrix} representation \cite{Rupp2012,Hansen2013}. Given a state vector $x$, the Coulomb matrix representation $C$ is a function of the pairwise distances $|p_k - p_{\ell}|$ and of the charge products $z_k z_{\ell}$
\begin{equation*}
C_{k,\ell} = \left\{
\begin{array}{ll}
\frac{1}{2}z_k^{2.4}, & k=\ell, \\[5pt]
\displaystyle\frac{z_kz_{\ell}}{|p_k-p_{\ell}|}, & k \neq \ell.
\end{array}
\right.
\end{equation*}
This representation thus satisfies the isometry invariance and deformation stability properties. 
However, it is not invariant to the permutation of the atom indices, which is a priori 
arbitrary. Although \cite{Hansen2013} proposes many strategies to mitigate this problem, it remains a challenge to this day. The most successful strategy is to augment the data set by associating to each molecule several permutations of its Coulomb matrix. The final predicted energy is then the average of the predicted energy for each permutation. While this technique improves performance, the data augmentation can significantly increase the size of the data set. In the context of kernel ridge regression, which achieves some of the best reported numbers for Coulomb matrices, this means that the size of the kernel can be very large. Furthermore, the fact that the size of the Coulomb matrix depends on the number of atoms $K$ in the molecule is another limitation. To remedy that issue, a fixed maximum size is set \textit{a priori}, and small Coulomb matrices are padded with zeros on the remaining rows and columns. However, once the training phase is complete, this approach effectively sets an upper bound on the molecular size supported by the representation. Finally, while the Coulomb matrix representation features multiple molecular length scales, it treats them on an equal footing. In particular, it does not take advantage of the multiscale structure of the energy, that emphasizes some scales more than others. It is possible though that the highly nonlinear regressors that couple with Coulomb matrices make up for this shortcoming. Numerical results seem to confirm that intuition, as linear regression with Coulomb matrices is an order of magnitude worse.

\section{Energy regression from electronic densities}
\label{energy_regression}

Hohenberg and Kohn proved in \cite{Kohn1964} that the molecular energy $E$ can be written as a functional of the electron density $\rho(u) \geq 0$ which specifies the density of electronic charge at every point $u \in \R^3$. 
The minimization of $E(\rho)$ over a set of electron densities $\rho$ leads to the calculation of the ground state energy 
\begin{equation*}
f(x) = E(\rho_x) = \inf_{\rho} E(\rho)~.
\end{equation*}
Hohenberg and Kohn also proved that there is a one to one
mapping between $\rho_x (u)$, the minimizing density, and $x$. In this work, we consider neutral molecules for which the
total electronic charge is equal to the sum of the protonic charges $z_k$
so that $\int \rho_x (u) du = \sum_k z_k$.
Computing $\rho_x$ is as difficult as computing $E(\rho_x)$. The next section explains how
to replace $\rho_x$ by an approximate density while section \ref{sparse_regression_by_OLS}
describes a sparse linear regression of $E$.

\subsection{Electronic density approximations}

\begin{figure}[ht]
\begin{center}
\centerline{\includegraphics[width=\columnwidth]{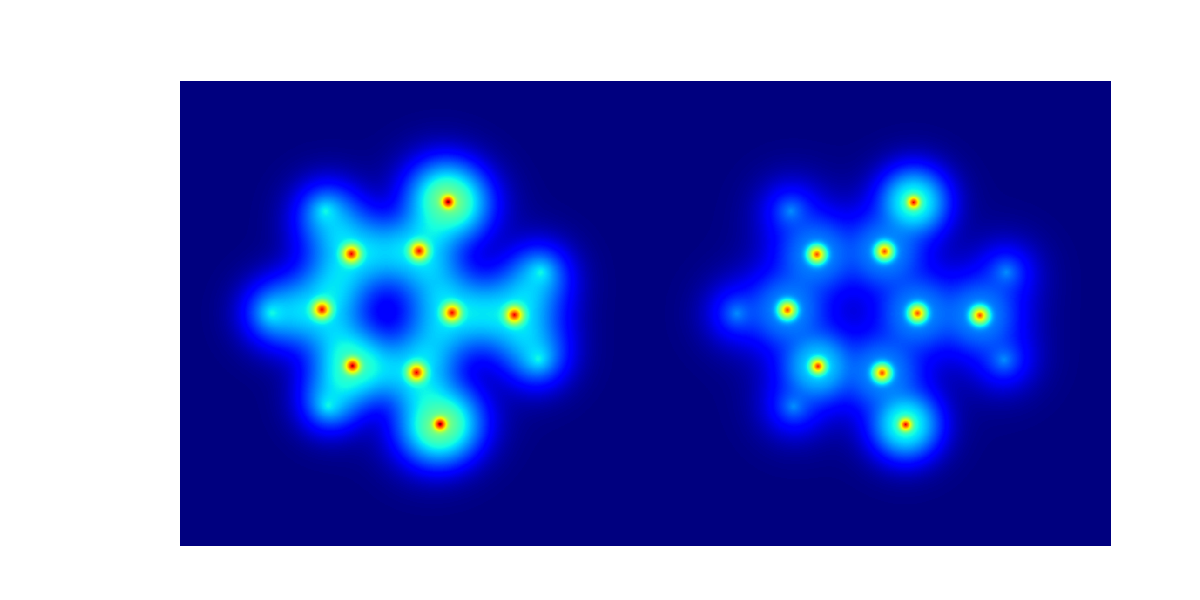}\hspace{0.35in}}
\caption{(left) Ground state electron density $\rho_x$ and (right): Approximate
electron density $\tilde \rho_x$ from \ref{model_densities}.}
\label{density}
\end{center}
\end{figure}

We use the spirit of the electron density approach by representing
$x$ by a crude approximate density $\tilde \rho_x$ of $\rho_x$ which also has a one to one
mapping with $x$ and satisfies $\int \tilde \rho_x (u) du = \sum_k z_k$. 
By construction, this approximate density is invariant to permutations of atom indices $k$ and its expression is given by:
\begin{equation}
\label{model_densities}
\tilde \rho_x (u) = \sum_{k=1}^{K} \rho_{\text{at}}^{a(k)}(u-p_k),
\end{equation}
which represents a simple linear superposition of isolated atomic densities. 
The notation $a(k)$ is a shorthand for the chemical nature of atom $k$ which determines its nuclear charge $z_k$, and hence which atomic density should be substituted. 
Isolated atomic densities are pre-computed from \href{http://en.wikipedia.org/wiki/Density_functional_theory}{Density Functional Theory} calculations for every distinct atomic species present in a molecular database.
The electron density model \eqref{model_densities}
only gives a crude approximation to $\rho_x$. It is a sum of
independent atomic contributions and hence does not model any chemical
effects like bond sharing. An example of an exact 
and approximate electron density is shown on Figure \ref{density}. The effect of bond sharing in $\rho_x$ appears as higher density "bridges" between atoms. These bridges are almost entirely absent in the case of $\tilde \rho_x$.

In this work, molecules are bi-dimensional. As a consequence, we
transform the three dimensional density in \eqref{model_densities}
into a two dimensional one, and use that later as our intermediate
representation. The dimensionality reduction in the density is
obtained by replacing each of the three dimensional atomic densities
with two dimensional ones. That last transformation is obtained by
"condensing" the three dimensional charge of a spherical shell of
radius $r$ and width $dr$ onto an annulus with the same radius and
width.

The resulting approximate density representation $\tilde \rho_x$ is
invariant to permutations of atom indices but it is not invariant to
isometries nor is it multiscale. The regression of the molecular energy $E(\rho_x)$ is therefore not computed from $\tilde \rho_x$ but from a representation 
$\Phi(\tilde \rho_x)$ which satisfies these properties. 
Section \ref{dictaser} will explain how to construct 
such representations $\Phi(\tilde \rho_x) = \{ \phi_k(\tilde \rho_x) \}_k$, while the
next section explains how to use them and approximate the energy $E(\rho_x)$ from a sparse linear regression calculated from a data set of training examples:
\begin{equation} \label{eqn: linear regression}
\tilde f(x) = \tilde E(\tilde \rho_x) = \sum_{k} w_k \phi_k(\tilde \rho_x )~.
\end{equation}

\subsection{Sparse Regression by Orthogonal Least Square}
\label{sparse_regression_by_OLS}

Given a training set of $N$ molecular state vectors and associated energies $\{x_i \, ,\, f(x_i) \}_{1 \leq i \leq N}$,
we explain how to compute the sparse regression in \eqref{eqn: linear regression}. 
To simplify notations, we shall write $\phi_k (\tilde \rho_x) = \phi_k (x)$.
A sparse $M$ term regression is obtained by selecting $M$ functionals $\{\phi_{k_m}\}_{1 \leq m
  \leq M}$ and computing an optimized linear regression
\begin{equation*}
\tilde f_M (x) = \sum_{m=1}^M w_m\, \phi_{k_m} (x),
\end{equation*}
which minimizes the quadratic error on training examples
\begin{equation}
\label{energy}
\sum_{i=1}^M \left|\sum_{m=1} w_m \phi_{k_m} (x_i) - f(x_i)\right|^2.
\end{equation}
These $M$ functionals are selected with a greedy orthogonal least square 
forward selection algorithm \cite{ChenOLS1991}.
The procedure selects and orthogonalizes each functional, one at a time. 

At the $m^{th}$ iteration it selects $\phi_{k_m}$, and a Gram-Schmidt orthogonalization
yields an orthonormalized $\phi_{k_m}^{\bot}$, 
which is uncorrelated relative to all previously selected functionals:
\begin{eqnarray*}
\forall \, n < m, \quad \sum_i \phi_{k_m}^{\bot} (x_i)\,\phi_{k_n}^{\bot} (x_i) = 0, \\
~~\mbox{and}~~\sum_i |\phi_{k_m}^{\bot} (x_i)|^2 = 1.
\end{eqnarray*}
The dictionary element $\phi_{k_m}^{\bot}$ is selected so that the linear regression of 
$f$ over $\{ \phi_{k_n}^{\bot} \}_{1 \leq n \leq m}$ minimizes the
quadratic error $\sum_i |\tilde f_m (x_i) - f(x_i)|^2$. 
This is equivalent to finding $\phi_{k_m}^{\bot}$ so that
$\sum_i f (x_i) \, \phi_{k_m}^{\bot} (x_i)$ is maximized. The
algorithm can be implemented with a $QR$ factorization, as described in
\cite{blumensathOLS2007}. The $M$-term regression can then be written as a function
of the original functionals $\phi_{k_m}$ but it is more easily expressed in
the orthogonalized Gram-Schmidt basis $\{ \phi_{k_m}^{\bot} \}_{m \leq M}$:
\begin{equation*}
\tilde f_M(x) = \sum_{m=1}^M w_{m}^{\bot} \phi_{k_m}^{\bot}(x)
\end{equation*}
where each coefficient is the correlation:
\[
 w_{m}^{\bot} = \sum_i f(x_i) \, \phi_{k_m}^{\bot}(x_i)~.
\]
The parameter $M$ is the dimension of the regression model. Increasing $M$ reduces
the bias error but also increases the variance error. The optimal $M$
results from a bias-variance trade-off. It is estimated with a cross validation over training examples. 

The bias error is the minimum approximation error of $f(x)$ from an $M$ term linear combination
of dictionary vectors $\{ \phi_k (x) = \phi_k(\tilde \rho_x) \}_k$. 
This error is small if $f(x)$ is well approximated by an element of a space spanned by $M$ terms of the dictionary.
This can be true only if the functionals $\phi_k(\tilde \rho_x)$ have the same invariance and stability properties as $f(x)$. The next section explains how to construct such a dictionary.

\section{Invariant representations}
\label{dictaser}

The central issue is to define a dictionary 
$\Phi(\tilde \rho_x) = \{ \phi_k (\tilde \rho_x)\}_k$ which is invariant to 
isometries, stable to deformations, multiscale, and sufficiently rich to perform
an accurate regression of the energy $f(x) = E(\rho_x)$. 
The numerical study is performed over
planar molecules. We thus restrict $\tilde \rho_x (u)$ over the molecular plane
$u \in \R^2$ and normalize it so that $\int_{\R^2} \tilde \rho_x(u) du = \sum_k z_k$. 
Invariance to isometries and stability to deformations is therefore
defined in $\R^2$.
To understand the challenge of defining
such a representation we begin by defining Fourier and wavelet representations and explain the
limitations of these two approaches. We then motivate the use of the Scattering representation introduced in \cite{Mallat2012}, to systematically construct stable invariants for regression. The extension to non-planar molecules in $\R^3$ involves no mathematical difficulty.

\subsection{Fourier Invariants}

Isometry invariant Fourier type representations based on the bispectrum and spherical harmonics are described in \cite{bartok:gaussAppPot2010, bartok:repChemEnviron2013}, and are used to regress potential energy surfaces for the dynamics of single molecules. We present a Fourier representation of an arbitrary function $\rho (u)$, which is invariant to linear isometric transformations of $u$, and discuss some of the limitations of Fourier based representations. 

Let $\hat{\rho}$ denote the Fourier transform of $\rho$:
\begin{equation*}
\hat{\rho}(\omega) = \int_{\R^2} \rho(u) e^{-iu \cdot \omega} \, du.
\end{equation*}
The modulus of the Fourier transform is a translation invariant representation of $\rho$. To add rotation invariance, we take $L^p$ averages over circles of radii $\gamma \in \R^+$:
\begin{equation} \label{eqn: fourier functionals}
\phi_{\gamma,p}^p(\rho) = \int_{|\omega| = \gamma}
|\hat{\rho}(\omega)|^p \, d\omega.
\end{equation}
In order to obtain a finite dictionary, we evenly sample the radii up to a maximum radius $R$, and we build the dictionary out of $L^1$ and $L^2$ terms for $p=1$ and $p=2$:
\begin{equation*}
\Phi_F(\rho) = \{ \phi_0(\rho), \phi_{\gamma,1}(\rho), \phi_{\gamma,1}^2(\rho), \phi_{\gamma,2}^2(\rho) \}_{\gamma}.
\end{equation*}
We use $L^1$ and $L^2$ terms to capture linear and quadratic dependencies in the energy functional.
In particular, one can prove that an important part of the exact Density Functional $E(\rho)$, 
namely the Hartree electron-electron repulsion functional, 
is a weighted sum of $L^2$ Fourier terms \eqref{eqn: fourier functionals} for $p=2$. 
These results extend for non-planar molecules by replacing the integrations in $\R^2$
by integrations in $\R^3$.

Numerical results from Table \ref{fig:MAE} provide the regression error obtained over a
Fourier dictionary $\Phi_F (\tilde \rho_x)$ computed from the atomic density approximation $\tilde \rho_x$ in \eqref{model_densities}.
The Fourier dictionary can regress atomization energies to nearly $10$ kcal/mol on average, 
which is a relatively large error. This indicates that the ground state energy is not well 
approximated in the functional space generated by the Fourier invariants. The main reason 
is that the Fourier representation $\Phi_F (\tilde \rho_x)$ is not stable to molecular deformations.
If $x$ and hence $\tilde \rho_x$ is deformed then high frequency terms will experience a large change in value. 
The same issue of instability to deformations appear for bispectrum representations. This
instability can be reduced by replacing the Fourier transform by a windowed
Fourier transform as in \cite{bartok:repChemEnviron2013}) for the
regression of potential energy surfaces. However, such a representation then only 
captures localized interactions within the window size, which must not be too large to avoid
deformation instabilities. Long range interactions, which represent a
non-negligible part of the energy of many aromatic organic molecules are not captured by such representations. Stability to deformations and capturing
short range and long range interactions requires a multiscale representation, which
motivates the use of a wavelet representation.

\subsection{Wavelet Invariants} \label{sec: wavelet invariants}

A wavelet is a complex valued function $\psi$ having zero average. We
suppose, additionally, that $\psi$ decays exponentially away from zero
and that $\psi(-u) = \psi^{\ast}(u)$, where $\psi^{\ast}$ denotes the
complex conjugate. We utilize Morlet wavelets, defined as:
\begin{equation*}
\psi(u) = \exp \left( - \frac{u_1^2 + u_2^2/\zeta^2}{2} \right)(\exp (i \xi u_1) - C).
\end{equation*}
The slant $\zeta > 1$ yields an anisotropic Gaussian which controls
the angular sensitivity of $\psi$. The Fourier transform of $\psi$ is
concentrated in a frequency domain centered at $(\xi,0)$, while the
constant $C$ is set so that $\int \psi = 0$.

Normally, dyadic wavelets are dilated by scales $2^j$ for $j \in
\Z$. We introduce a scale oversampling by a factor $2$ and dilate the
wavelet at scales $2^{j/2}$. In the following, we are concentrating on
wavelets defined in two dimensions $u \in \R^2$. Two dimensional
wavelets are rotated by $r_\theta$ for an angle $\theta \in
[0,2\pi)$. A dilated and rotated wavelet is then indexed as:
\begin{equation*}
\psi_{j,\theta}(u) = 2^{-2j/2} \psi (2^{-j/2}r_{-\theta}u),
\end{equation*}
and the wavelet transform is defined by
\begin{equation*}
\rho \mapsto \{ \rho \star \psi_{j,\theta}(u) \}_{j,\theta,u}.
\end{equation*}
From the wavelet transform we derive isometry invariant functionals at
different scales by averaging over translations $u \in \R^2$ and
rotations $\theta \in [0,2\pi)$:
\begin{equation} \label{eqn: wavelet functionals}
\phi_{j,p}^p(\rho) = \int_{\R^2} \int_0^{2\pi} | \rho \star
\psi_{j,\theta}(u)|^p \, d\theta \, du.
\end{equation}
As with the Fourier dictionary, the wavelet dictionary is made finite
by utilizing a finite range of scales and various $L^1$ and $L^2$
functionals:
\begin{equation*}
\Phi_W(\rho) = \{ \phi_0(\rho), \phi_{j,1}(\rho), \phi_{j,1}^2(\rho), \phi_{j,2}^2(\rho) \}_j.
\end{equation*}

The Fourier functionals \eqref{eqn: fourier functionals} integrate the frequency energy of $\rho$ over circles of radii $\gamma$. The wavelet functionals, however, take advantage of the multiscale structure of the energy and integrate the frequency energy of $\rho$ over annuli of bandwidth $2^{j/2}$. Furthermore, as shown in \cite{Mallat2012}, the wavelet functionals linearize the action of diffeomorphisms on $\rho$. Numerical results from Table \ref{fig:MAE} indicate that the wavelet dictionary requires significantly fewer coefficients to achieve a comparable accuracy as the Fourier dictionary. However, the minimum average error remains similar, as the two dictionaries contain similar frequency information on the density $\rho$. Thus, while the wavelet dictionary satisfies all of the properties of Section \ref{desirable_properties_of_representations}, it is not a rich enough dictionary to model the energy $E$ to state-of-the-art accuracy. 
This final limitation is resolved through the scattering transform. 
These results extend in $\R^3$ with a wavelet transform over $\R^3$.

\subsection{Scattering}

The scattering transform augments the wavelet dictionary by providing
additional multiscale invariants through iterated wavelet
transforms. This architecture is a type of deep
convolutional network, with variations of it applied
successfully in computer vision for texture classification
\cite{mallat:rotoScat2013,MallatBruna2013} as well as object
classification \cite{oyallon:scatObjectClass2014}.

Deep convolutional networks cascade linear and nonlinear operations
through multilayer architectures \cite{bengio2013representation}.
In the first layer, features from a two
dimensional function $\rho$ are extracted via a collection of functionals
$\{ h_k \}_k$. These functionals apply a localized linear filter $\LL_k$
across the function $\rho$ via convolution, followed by a
nonlinear function $F$ that also downsamples the signal,
\begin{equation*}
h_k(\rho) = F(\rho \star \LL_k).
\end{equation*}
The linear filters $\LL_k$ are learned from the training set via
back-propagation. The nonlinear functions may be sigmoids, rectifiers
or absolute values, followed by a pooling operator.
Subsequent layers convolve linear filters both spatially and
over the collection of functionals $\{ h_k \}_k$ from the previous
layer, thus combining information across filters. 

Wavelets $\{\psi_{j_1,\theta_1}\}_{j_1,\theta_1}$ are predefined linear
filters that capture information at scale $2^{j_1/2}$ and in the direction
$\theta_1$. The complex modulus is a pointwise nonlinear function which, when
applied after the wavelet transform, yields functionals $\{ |\rho \star
\psi_{j_1,\theta_1}| \}_{j_1,\theta_1}$ analogous to $\{ h_k \}_k$. These
functionals are invariant to translation up to scale $2^{j_1/2}$. Since the
energy is globally invariant to isometries, the wavelet
invariant functionals $\{ \phi_{j_1,p} \}_{j_1}$ of \eqref{eqn: wavelet
  functionals} are derived from global $L^p$ averages (pooling) of $\{
| \rho \star \psi_{j_1,\theta_1}(u) | \}_{j_1,\theta_1}$ over translations $u$
and rotations $\theta_1$. However, this integration removes the
variation of $|\rho \star \psi_{j_1,\theta_1}(u)|$ along $(u,\theta_1)$,
thus discarding a considerable amount of information. In order to
recover some of this lost information we apply a second layer of
wavelet transforms to the collection of functions $\{|\rho \star
\psi_{j_1,\theta_1}(u)|\}_{j_1,\theta_1}$.

For fixed $(j_1,\theta_1)$, the function $u \mapsto |\rho \star
\psi_{j_1,\theta_1}(u)|$ varies at scales bigger than
$2^{j_1/2}$. Translation information at scale $2^{j_1/2}$ and angle
$\theta_1$ is recovered by applying a second
spatial wavelet transform, with the same Morlet wavelet, for scales
$j_2 > j_1$ and angles $\theta_2$, yielding $\{ |\rho \star \psi_{j_1,\theta_1}| \star
\psi_{j_2,\theta_2} \}_{j_2,\theta_2}$. Applying a second wavelet
transform to each of the functions from the first layer wavelet
transform gives the following collection of second layer spatial functions:
\begin{equation} \label{eqn: spatial part of 2nd layer}
\{ | \rho \star \psi_{j_1,\theta_1} | \star \psi_{j_2,\theta_2}(u) \}_{j_1,\theta_1,j_2,\theta_2,u}.
\end{equation}

The collection \eqref{eqn: spatial part of 2nd layer}, while
recovering lost spatial information, does not recover the angular
variability of the first layer considered as a function $\theta_1 \mapsto
|\rho \star \psi_{j_1,\theta_1}(u)|$ for fixed $j_1$ and $u$. For scales
$2^{j_1/2}$ on the order of the distance between neighboring atoms, 
the behavior of these functions reflects the orientation of atomic
bonds and hence bond angles. The variability over $\theta_1$ for larger
spatial scales indicates global geometric structure, such as the
orientation of sub-molecules. This rotation information is recovered
by applying a wavelet transform over the angles $[0,2\pi)$ along the
rotation variable $\theta_1$. The wavelet transform is defined in terms
of circular convolution:
\begin{equation*}
g_1 \circledast g_2 (\theta) = \int_0^{2\pi} g_1(\theta') g_2(\theta-\theta') \, d\theta'.
\end{equation*}
Periodic dilated wavelets are defined over $[0,2\pi)$ by periodizing a
one dimensional dilated Morlet wavelet $\psi^{\mathrm{1D}}: \R \rightarrow \C$,
\begin{equation*}
\overline{\psi}_{\ell_2} (\theta) = 2^{-\ell_2} \sum_{k \in \Z} \psi^{\mathrm{1D}}(
2^{-\ell_2}\theta - 2\pi k).
\end{equation*}
The resulting angular part of the second layer transform
is then:
\begin{equation} \label{eqn: angular part of 2nd layer}
\{ | \rho \star \psi_{j_1,\cdot}(u) | \circledast
\overline{\psi}_{\ell_2}(\theta_1)  \}_{j_1,\theta_1,j_2,\theta_2,u}.
\end{equation}
Combining the second layer spatial transform \eqref{eqn: spatial part
  of 2nd layer} and the second layer angular transform \eqref{eqn:
  angular part of 2nd layer}, and applying the complex modulus, yields
the following collection of functions:
\begin{equation} \label{eqn: 2nd layer U}
\rho \mapsto \{ || \rho \star \psi_{j_1,\cdot} | \star \psi_{j_2,\theta_2}(u) \circledast \overline{\psi}_{\ell_2}(\theta_1)| \}_{j_1,\theta_1,j_2,\theta_2,\ell_2,u}.
\end{equation}
As in the wavelet dictionary, isometry invariant functionals are
derived from these second layer wavelet transforms via integration
over translations and rotations. Note, however, that a rotation of
$\rho$ propagates through the layers of the scattering
transform. Indeed, if $\rho_{\theta}(u) = \rho(r_{-\theta} u)$ is a
rotation of $\rho$ by angle $\theta$, then
\begin{align*}
||\rho_{\theta} \star \psi_{j_1, \cdot}| &\star
\psi_{j_2,\theta_2}(u) \circledast \overline{\psi}_{\ell_2}(\theta_1)|
  = \\
&||\rho \star \psi_{j_1, \cdot}| \star
\psi_{j_2,\theta_2-\theta}(r_{-\theta} u) \circledast \overline{\psi}_{\ell_2}(\theta_1-\theta)|
\end{align*}
Thus both angular variables $\theta_1$ and $\theta_2$ are rotationally
covariant. However, an orthogonal change of coordinates $(\theta_1,
\theta_2) \mapsto (\alpha, \beta)$, with:
\begin{equation*}
\alpha = \frac{\theta_1-\theta_2}{2}, \qquad \beta = \frac{\theta_1 + \theta_2}{2},
\end{equation*}
yields one rotationally invariant variable $\alpha$ and one
rotationally covariant variable $\beta$. Thus isometry invariant
scattering functionals can be derived from \eqref{eqn: 2nd layer U} by
integrating over $(u, \beta) \in \R^2 \times [0,2\pi)$:
\begin{align*}
&\phi_{j_1,\lambda_2,p}^p (\rho) = \\
&\int_{\R^2} \int_0^{2\pi}|| \rho \star \psi_{j_1,\cdot} |
  \star \psi_{j_2,\beta-\alpha}(u) \circledast
  \overline{\psi}_{\ell_2}(\alpha+\beta) |^p \, d\beta \, du,
\end{align*}
where $\lambda_2 = (j_2,\alpha,\ell_2)$ encodes the parameters of the second layer. A finite scattering dictionary is obtained by taking a finite number of scales $j_1, j_2, \ell_2$ as well as a finite number of angles $\alpha$, for a mixture of $L^1$ and $L^2$ functionals:
\begin{align*}
\Phi_S(\rho) = \{ &\phi_0(\rho), \phi_{j_1,1}(\rho), \phi_{j_1,1}^2(\rho), \phi_{j_1,2}^2(\rho), \ldots \\ &\ldots \phi_{j_1,\lambda_2,1}(\rho), \phi_{j_1,\lambda_2,1}^2(\rho), \phi_{j_1,\lambda_2,2}^2(\rho) \}_{j_1,\lambda_2}.
\end{align*}

The second layer of the scattering transform greatly expands the wavelet dictionary, but still satisfies the four properties of Section \ref{desirable_properties_of_representations}. Numerical experiments show (see Table \ref{fig:MAE}) that the average regression error over the scattering dictionary is greatly reduced, to approximately $1.8$ kcal/mol, thus indicating that the second layer functionals dramatically increase the ability of the dictionary to model the full variability of the energy.
The extension of a scattering transform in $\R^3$ is done with three-dimensional
spatial wavelet transforms, indexed by a direction $\vec \theta$ which belongs to the
two-dimensional sphere in $\R^3$. The second order scattering coefficients are then
calculated with a wavelet transform along $\vec \theta$, and thus 
on the two-dimensional sphere in $\R^3$ \cite{starksphere}.

\section{Numerical experiments}
\label{sec: numerical experiments}

We compare the performance of Coulomb matrix representations with Fourier, wavelet
and Scattering representations on two databases of \textit{planar} organic molecules.
Molecular atomization energies from these databases were computed using the PBE0 hybrid density functional \cite{Adamo1999}. The first database includes $454$ nearly planar molecules among the $7165$ molecules of the QM7 molecular database \cite{Rupp2012}. We also created a second database of $4357$ strictly planar molecules, which we denote QM2D.
We produced this new database with a similar procedure as 
the one outlined in \cite{Rupp2012}.
Both databases consist of a set of organic molecules composed of hydrogen, carbon,
nitrogen, oxygen, sulfur and additionally chlorine in the case of QM2D. The molecules featured in these sets cover a large spectrum of representative organic groups typically found in Chemical Compound Space \cite{Rupp2012}. Particular care was taken in producing well-balanced folds used in cross validation assessments. A proper partitioning of the data points among folds was indeed outlined in \cite{Hansen2013} as critical to ensure low variance in test errors.

\subsection{Coulomb matrix baseline and error metrics} \label{sec: numerical
  results and comparisons}

In the case of the Coulomb matrix based regression we used the best performing representation assigning eight randomly sorted Coulomb matrices per molecule as described in \cite{Hansen2013}. The width $\sigma$ of the Laplace kernel and the ridge parameter $\lambda$ were selected following the algorithm described in the same paper.
The algorithm was validated by recovering the published accuracy over the full QM7
database which contains $7165$ molecules. In our experiments, we restrict 
the database size to only 454 planar molecules so the regression error is larger. The same methodology is then followed to compute the optimal Coulomb matrix based regression on QM2D.

To evaluate the precision of each regression algorithm, each database is broken into five representative folds,
and all tests are performed using five fold cross validation in which we reserve four folds for training, and the fifth fold for testing. This results in a regressed energy for each molecule in the database. We report the Mean Absolute Error (MAE or $\ell^1$) over each database along with the Root Mean-Square Error (RMSE or $\ell^2$), which is the square root of the average squared error.

\begin{table*}[t]
\footnotesize
\vskip 0.15in
\renewcommand{\arraystretch}{2.5}
\center
\caption{Error in kcal/mol of regressed quantum molecular energies using different molecular representations (vertically) and different error measures (horizontally), on two databases of planar organic molecules: left and right parts of the table.}
\vskip 0.15in
\begin{tabular}{|c|c||c|c|c||c|c|c|}
\hline
\multicolumn{2}{|c||}{}& \multicolumn{3}{|c||}{454 2D molecules from QM7}
  & \multicolumn{3}{|c|}{4357 molecules in QM2D} \\
\cline{3-8}
\multicolumn{2}{|c||}{}& $M$ & $\ell^1$: MAE & $\ell^2$: RMSE  & $M$ & $\ell^1$: MAE & $\ell^2$: RMSE \\
\hline
\hline
\multicolumn{2}{|c||}{Coulomb} & N/A & 7.0 & 20.5 & N/A & 2.4 & 5.8 \\
\hline
\multicolumn{2}{|c||}{Fourier} & 62 & 11.9 & 16.1 & 198 & 11.1 & 16.7 \\
\hline
\multicolumn{2}{|c||}{Wavelet} & 42 & 11.1 & 15.5 & 59 & 11.1 & 14.2 \\
\hline
\multicolumn{2}{|c||}{Scattering} & 74 & {\bf 6.9} & {\bf 9.0} & 591 & {\bf 1.8} & {\bf 2.7} \\ 
\hline
\multicolumn{2}{|c||}{Chemical Accuracy} & \multicolumn{6}{|c|}{1.0} \\
\hline
\end{tabular}
\label{fig:MAE}
\end{table*}

\subsection{Dictionary Implementations and Sizes}

The number of elements in
the Fourier, wavelet, and scattering dictionaries are very different and respectively
equal to $1537$, $61$ and $11071$ in numerical computations. This section explains
the implementation of these dictionaries.
Molecular configurations are centered at zero, and the two dimensional electron density $\tilde\rho_x$ is restricted to a box $[-a,a]^2$. The parameter $a$ is chosen so that the density decays to nearly zero at the boundary (in our experiments, $a=11$ angstroms). The box is then sampled with $2^J \times 2^J$ evenly spaced grid points, for some resolution $J$; in practice $J=10$ for the Fourier and wavelet dictionaries, and $J=9$ for the scattering dictionary. The grid naturally leads us to a discretized version of $\tilde\rho_x$.

For the Fourier representation, discrete Fourier transforms are
computed with a two dimensional Fast Fourier Transform
(FFT). Integration over circles of radii $\gamma$ is approximated with
finite sums over discretized circular contours, which are approximated
using the original $2^J \times 2^J$ spatial sampling. This yields $2^{J-1}$ functionals for a fixed
$L^p$ average, resulting in a dictionary of size $1 + 3 \cdot
2^{J-1}$. For $J=10$, we have $1537$ dictionary elements. 

The wavelet parameters are set according to the following
specifications. The minimum scale is $j_{\min} = 0$ and the maximum
scale is $j_{\max} = J - 1/2$, resulting in a total of $2J$ scales for
the wavelet transform. $L$ angles are evenly sampled from $[0,\pi)$
according to $\theta = k\pi/L$ for $k=0,\ldots,L-1$, which is
equivalent to evenly sampling $2L$ angles over $[0,2\pi)$ since the
Morlet wavelet is symmetric relative to both the $x$ and $y$ axes. In
practice we take $L=8$. The slant is $\zeta = 1/2$, and the central
frequency is fixed at $\xi = 3\pi/4$. Given a fixed slant $\zeta$ and
central frequency $\xi$, increasing $J$ and $L$ should not decrease
the accuracy of the regression. As a consequence, we fixed the values
for these parameters when the regression results appeared to
plateau. Wavelet transform convolutions are computed as
multiplications in frequency space by utilizing FFTs and inverse
FFTs. The modulus of the output functions from these convolutions are
then averaged over the discrete spatial grid and the discrete sampling
of $[0,\pi)$ to obtain the wavelet dictionary functionals. This
results in a wavelet dictionary of size $1 + 6J$. For $J=10$, this
yields $61$ dictionary elements

For the second layer of the scattering transform, scales $j_2$ are
computed for $j_2 > j_1$, resulting in $2J(2J-1)/2$ pairs
$(j_1,j_2)$. Angles $\theta_2$ are discretized with $L$ samples, and
the wavelet transform over $[0,\pi)$ is computed for dyadic scales
$2^{\ell_2}$ with $\ell_2=0,\ldots,\log_2L$. The total size of the
scattering dictionary is then $1 + 6J + 3 (2J (2J-1) L \log_2 L) /
2$. For $J=9$ and $L=8$, the scattering dictionary contains $11071$
functionals.

\section{Results and discussion}
\label{results}

The results from our numerical experiments are summarized in Table \ref{fig:MAE}. 
The Scattering representation computed from the atomic density model offers state-of-the-art accuracy. Over the larger database, it has a 25\% improvement in Mean Absolute Error over the best Coulomb matrix based regression technique. This improvement 
goes up to $53\%$ in Root Mean Square Error.
The larger improvement of the RMSE is due to the fact that a scattering regression has smaller error outliers.

\begin{figure}[ht]
\begin{center}
\centerline{\includegraphics[width=\columnwidth]{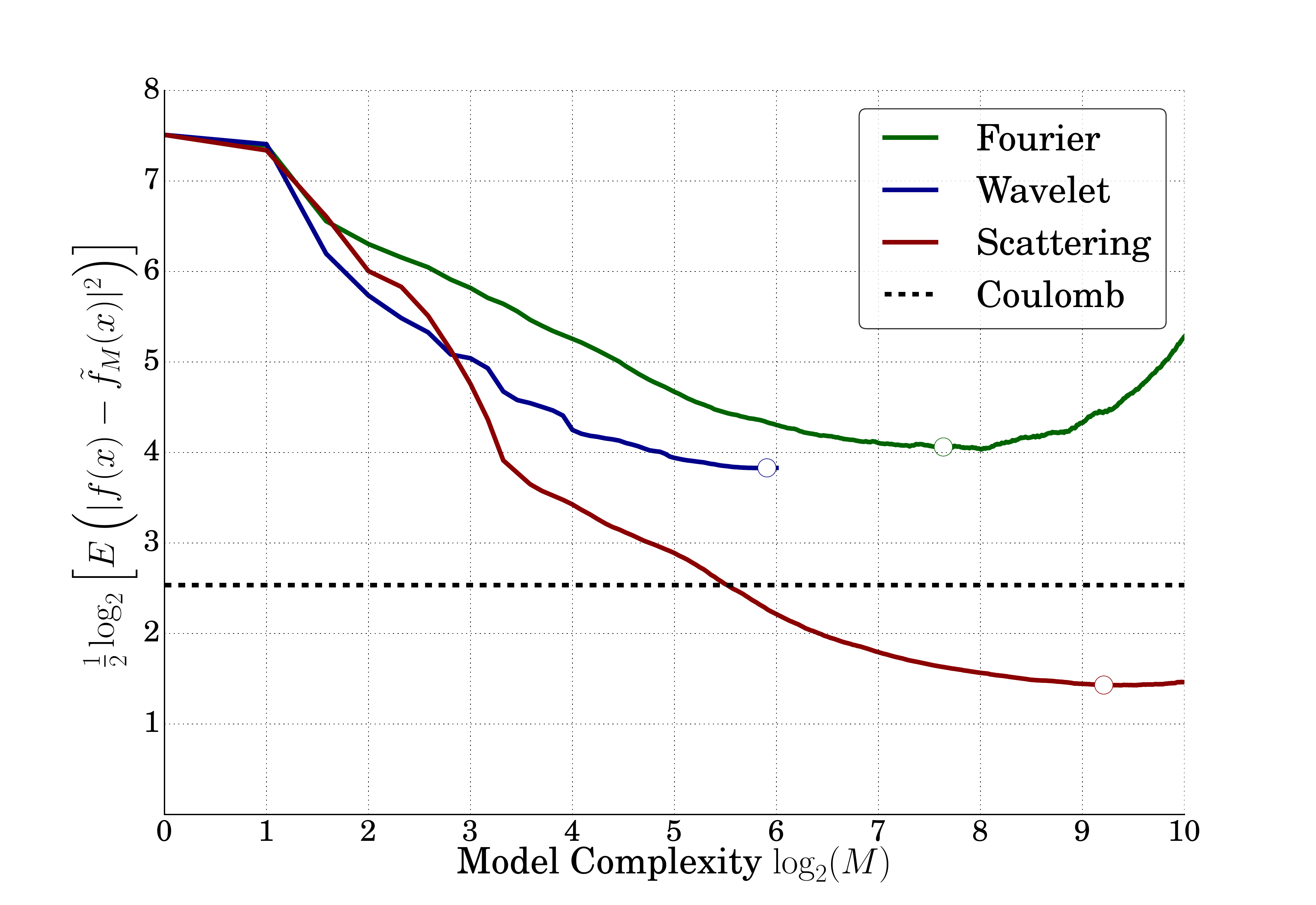}}
\caption{Decay of the log RMSE error $\frac{1}{2}\log_2\left[E\left(|f(x)-\tilde{f}_M(x)|^2\right)\right]$ over the
larger database of 4357 molecules, as a function of $\log_2(M)$ in
the Fourier (green), Wavelet (blue) and Scattering (red) regressions.
The dotted line gives the Coulomb regression error for reference.}
\label{bias_variance}
\end{center}
\end{figure}

Table \ref{fig:MAE} shows that the error of Fourier and wavelet regression 
are of the same order although the Fourier dictionary has $1537$ elements and the
wavelet dictionary has only $61$. 
Figure \ref{bias_variance} gives the decay of these errors as a function of $M$.
This exepected error is computed on testing molecules. 
The circles on the plot give the estimated value of $M$ which yield a minimum regression error
by cross-validation over the training set (reported in Table \ref{fig:MAE}).
Although the Fourier and wavelet regressions reach nearly the same minimum error, 
the decay is much faster for wavelets.
When going from the smaller to the larger database, the minimum error of the Fourier and wavelet regressions remain nearly the same. This shows that the bias error due to 
the inability of these dictionaries to precisely regress $f(x)$ is dominating the
variance error corresponding to errors on the regression coefficients. 
The Coulomb and Scattering representations on the other hand, achieve much smaller bias errors on the larger database.

The number of terms of the scattering regression is $M = 591$ on the larger database,
although the dictionary size is $11071$. A very small proportion of scattering
invariants are therefore selected to perform this regression. The chosen scattering coefficients used for the regression are coefficients corresponding to scales which fall between the minimum and maximum pairwise distances between atoms in the molecular database. These selected coefficients are thus adapted to the molecular geometries.

\section{Conclusion}
\label{conclusion}

This paper introduced a novel intermediate molecular representation through the use of a model electron density. The regression is performed on a scattering 
transform applied to a model density built from a linear superposition of atomic densities. This transform is well adapted to quantum energy regressions because it 
is invariant to the permutation of atom indices, to isometric transformations, it
is stable to deformations, and it separates multiscale interactions. It is computed
with a cascade of wavelet convolutions and modulus non-linearities, as a deep convolutional network. 
State-of-the-art regression accuracy is obtained over two databases of 
two-dimensional organic molecules, with a relatively small number of scattering
vectors. Understanding the relation between the choice of scattering coefficients and
the physical and chemical properties of these molecules is an important issue.

Numerical applications have been carried out over planar molecules,
which allows one to restrict the electronic density to the molecular
plane, and thus compute a two-dimensional scattering transform. A scattering transform is similarly defined in three
dimensions, with the same invariance and stability properties.
It involves computing a wavelet transform on the two-dimensional
sphere $S^2$ in $\R^3$ \cite{starksphere} as opposed to the circle $S^1$. 
It entails no mathematical difficulty, but requires appropriate software implementations which are being carried out.

Energy regressions can also provide estimations of forces through differentiations with respect to atomic positions. Scattering functions are differentiable
and their differential can be computed analytically. However, the precision of such
estimations remain to be established.

\section*{Acknowledgments} 

This work was supported by the ERC grant InvariantClass 320959.

\bibliography{icml2015paper}
\bibliographystyle{icml2015paper}

\end{document}